\documentclass[letterpaper, 10 pt, conference]{ieeeconf}
\IEEEoverridecommandlockouts    
\usepackage{graphics}           
\usepackage{times}              
\usepackage{amsmath}            
\usepackage{amssymb}            
\usepackage{graphicx}
\usepackage{algorithm}
\usepackage[noend]{algpseudocode}
\usepackage{booktabs}

\usepackage[font=small]{caption}

\def\secref#1{Sec.~\ref{#1}}
\def\figref#1{Fig.~\ref{#1}}
\def\tabref#1{Tab.~\ref{#1}}
\def\eqref#1{Eq.~(\ref{#1})}

\makeatletter
\usepackage{xspace}
\DeclareRobustCommand\onedot{\futurelet\@let@token\@onedot}
\def\@onedot{\ifx\@let@token.\else.\null\fi\xspace}

\def\etal{{et al}\onedot}
\makeatother

\usepackage{array}
\newcolumntype{L}[1]{>{\raggedright\let\newline\\\arraybackslash\hspace{0pt}}m{#1}}
\newcolumntype{C}[1]{>{\centering\let\newline\\\arraybackslash\hspace{0pt}}m{#1}}
\newcolumntype{R}[1]{>{\raggedleft\let\newline\\\arraybackslash\hspace{0pt}}m{#1}}

\newcommand{\abs}[1]{|#1|}

\usepackage{color} 
\usepackage{subfigure}
\usepackage{pifont} 
\usepackage{commath}
\newcommand{\cmark}{\ding{51}}%

\usepackage{hyperref}

\title{\LARGE \bf Simple But Effective Redundant Odometry for Autonomous Vehicles}

\author{Andrzej Reinke \and  Xieyuanli Chen \and Cyrill Stachniss
	\thanks{All authors are with the University of Bonn, Germany. }%
	\thanks{This work has partially been funded by the Deutsche Forschungsgemeinschaft (DFG, German Research Foundation) under Germany's Excellence Strategy, EXC-2070 - 390732324 (PhenoRob) and by the European Union’s Horizon 2020 research and innovation programme under grant agreement No~101017008~(Harmony).
	}
}

\begin{document}
	\maketitle
	\thispagestyle{empty}
	\pagestyle{empty}

\begin{abstract}
Robust and reliable ego-motion is a key component of most autonomous mobile systems. 
Many odometry estimation methods have been developed using different sensors such as cameras or LiDARs.
In this work, we present a resilient approach that exploits the redundancy of multiple odometry algorithms using a 3D LiDAR scanner and a monocular camera to provide reliable state estimation for autonomous vehicles.
Our system utilizes a stack of odometry algorithms that run in parallel. It chooses from them the most promising pose estimation considering sanity checks using dynamic and kinematic constraints of the vehicle as well as a score computed between the current LiDAR scan and a locally built point cloud map.
In this way, our method can exploit the advantages of different existing ego-motion estimating approaches.
We evaluate our method on the KITTI Odometry dataset. The experimental results suggest that our approach is resilient to failure cases and achieves an overall better performance than individual odometry methods employed by our system.
\end{abstract}

\section{Introduction}
\label{sec:intro}
The ability to estimate the ego-motion, also called odometry, is a vital part of most autonomous mobile systems.
Nowadays, autonomous vehicles are typically equipped with multiple sensors, such as cameras or LiDARs, as different sensing modalities have individual strengths.
Therefore, a large number of algorithms~\cite{zhang2015icra-voam, huang2018iros, huang2019icra} have been proposed exploiting both, visual and LiDAR information to explore the advantages of both and compensate for the drawbacks of the other.
In most cases, researchers focus on achieving better odometry results by designing a potentially complex multi-sensor fusion algorithm. 
Even if different pose estimation systems' performances are remarkable within specific datasets, such algorithms may not always operate reliably under all conditions, setups, and application domains, as we will illustrate below in~\secref{sec:robustness}. 
Thus, failure detection and recovery steps are useful in real-world applications.
We investigate in this paper a strategy that runs multiple odometry approaches in parallel and then selects the method that appears to be the best one. Thus, newly proposed individual methods can easily be added to our approach.

Few studies are investigating redundant odometry methods running in parallel on autonomous mobile systems. One example is the work by Luo \etal~\cite{luo2002sensors}, who propose to estimate odometry as an average of parallel odometry pipelines.
This approach, however, is vulnerable since already one gross estimation error of one method will affect the overall performance. 
Santamaria-Navarro \etal~\cite{santamarianavarro2020arxiv} propose another solution, which runs several methods in parallel. 
It first sticks to one odometry method and only falls back to an alternative odometry pipeline if the initial method's divergence is detected.
While this method is straightforward to realize, it does not utilize to its full extent the potential of all algorithms that run in parallel. 

\begin{figure}[t]
	\centering
	\includegraphics[width=0.95\linewidth]{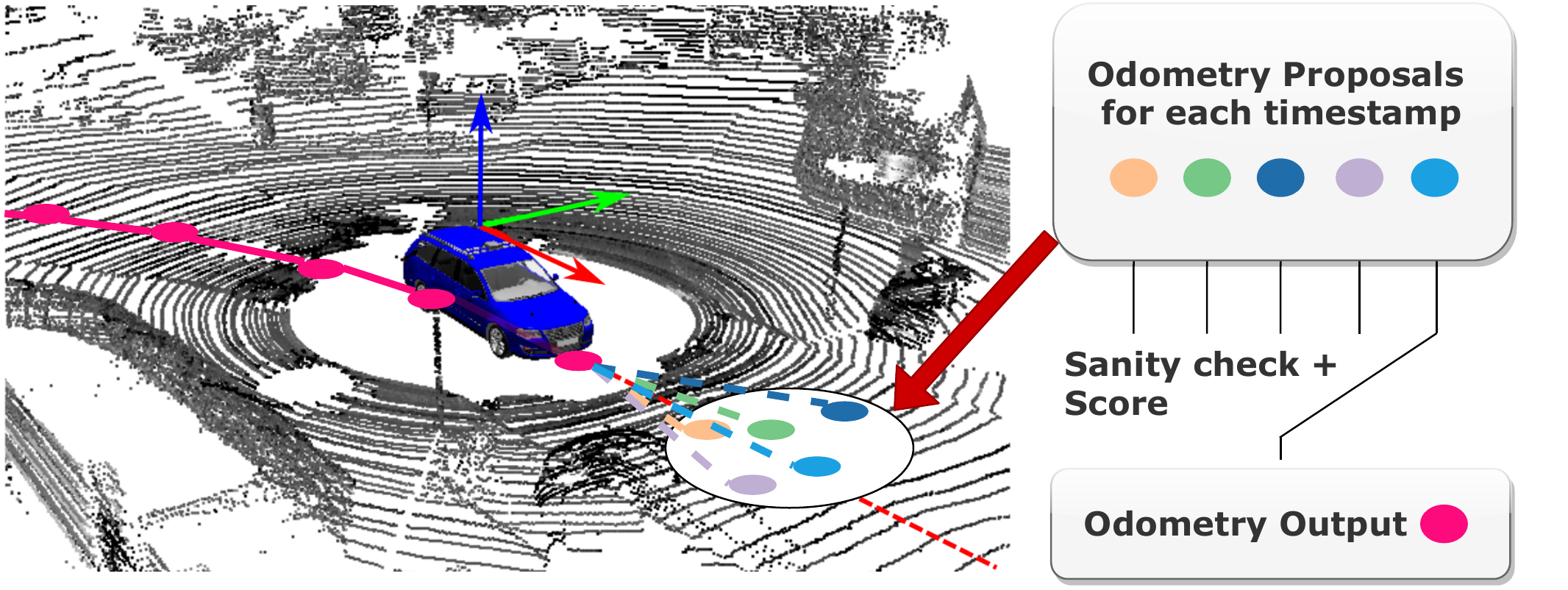}
	\caption{Our redundant odometry system employs multiple odometry algorithms to provide proposals shown as differently colored dots for the each timestamp, exploiting LiDAR and camera data. Our approach uses sanity checks and a Chamfer distance-based score to choose the most promising pose estimation from all candidates.}
	\label{fig:motivation}

\vspace{-0.5cm}

\end{figure}	

In this work, we investigate the possibility of building a redundant odometry system with sanity checks, and that can deal with failure cases of different odometry methods. 
As sketched in~\figref{fig:motivation}, instead of sticking to a specific odometry method, the idea is to run various odometry algorithms in parallel and choose the most promising one. We target to avoid explicit recovery behaviors for individual odometry pipelines in case of failures but propose to switch dynamically between estimation algorithms and independently of previous decisions. 
Furthermore, we target a computationally light-weight and flexible system.

The main contribution of this paper is a redundant odometry system to provide a more resilient ego-motion estimation than the individual odometry methods that the system consist of. Instead of sticking to one specific odometry method, the proposed system runs multiple odometry algorithms in parallel and exploits LiDAR and camera data. 
It combines sanity checks exploiting dynamic and kinematic constraints of the vehicle and Chamfer distance-based scores between sensor data and a local map to select the most promising pose estimation among all the odometry candidates.
We make three key claims:
Our approach is able to 
(i)~use redundant odometry pipelines to provide better odometry results,
(ii)~generally yields the best results compared to the individual methods used, and, 	
(iii)~avoids several failure cases in different situations.
The paper and our experimental evaluation back up these claims. Furthermore, the source code of our approach is available at:\\
{\small{\url{https://github.com/PRBonn/MutiverseOdometry}}}

\section{Related Work}
\label{sec:related}
Odometry estimation is a widely investigated topic in robotics. 
It can be achieved using different types of sensors, including wheels~\cite{thrun2005probrobbook}, inertial sensing~\cite{groves2015aesm}, cameras~\cite{scaramuzza2011arxiv}, thermal cameras~\cite{delaune2019iros}, radars~\cite{cen2018icra}, LiDARs~\cite{rusinkiewicz2001dim}, sonars~\cite{bahr2009ijrr} as well as various combinations~\cite{geiger2013ijrr, zhang2014rss, zhang2015icra-voam, chen2019iros, behley2019iccv, milioto2019iros}.
Although much work has been done, only a small amount of research has been devoted to making a robust and redundant system to deal with failures of individual ego-motion estimation approaches.

Works show that redundancy can make a robotic system more robust.
Rollinson \etal~\cite{rollinson2013dscc} show a robust estimation of the configuration of an articulated robot that uses a large number of redundant proprioceptive sensors (encoders, gyros, accelerometers) embedded in an unscented Kalman filter. 
They formulate the state estimation problem to leverages the redundancy in the proprioceptive information provided by the robot's joint angle encoders and inertial sensors. 
Additionally, they introduce a novel outlier detector that can identify corrupted measurements by utilizing the Mahalanobis distance matrix.
Kim \etal~\cite{kim2019ac}  point out that redundant systems are important due to adversary attacks that corrupt measurement. 
They propose an observer for a nonlinear system to detect sensor attacks.

The importance of a redundant odometry system has also been shown on Mars Exploration Rovers ~\cite{maimone2007jfr}. 
It successfully demonstrated a switching behavior between the wheel, inertia, and visual odometry, where visual odometry was used as compensation for any unforeseen slip encountered during a drive.
Another redundant odometry system was proposed by Luo \etal~\cite{luo2002sensors} taking an average estimate over all odometry estimates that are currently in the active mode. Others use fallback solutions and execute homing behaviors~\cite{perea2016jras},  sampling-based motion proposals in a particle filter~\cite{stachniss2007iros}, or combinations of dense and feature-based methods~\cite{wurm2010ras}.
Carnevale \etal~\cite{carnevale2015ijars} propose a way to fuse the information coming from two typologies of redundant radio-frequency identification sensors with complementary characteristics using a Kalman filter framework. 
A switching criterion is embedded in the filtering approach where the observer uses uncertainties of the measurements and their average values to provide a switching behavior.
Recently, Ebadi \etal~\cite{ebadi2020icra} created an autonomous mapping and positioning system for the DARPA Subterranean Challenge. 
This work uses visual-inertial, LiDAR, and wheeled odometry as a front-end for a multi-robot mapping and positioning system to explore perceptually-degraded environments. 
They use the generalized iterative closest point (GICP) algorithm~\cite{segal2009rss} to estimate scan-to-scan and scan-to-map matching that provides the relative transformation between the consecutive timestamps.

The related approach to our work has recently been proposed by Santamaria-Navarro \etal~\cite{santamarianavarro2020arxiv} in the context of UAVs. 
Their system called heterogeneous redundant odometry (HeRO) evaluates the pose estimations from each pipeline by checking gaps and jumps within the transforms sequence. In this method, the different odometries run in parallel, and the potentially best estimation is used as the output of the system until the uncertainty of the estimation exceeds certain quality criteria. 
Then, the system falls back to another method while at the same time trying to reinitialize the former method.
Unlike HeRO, which treats each odometry pipeline as a separate entity with a separated history of transformations and transformed point clouds,
our method does not need any reinitialization procedure. It switches to an alternative method or fallbacks to a constant velocity model if all other methods fail. 
We focus on picking the best transform independently from the previous transforms and use a chain of transformation and transformed point clouds as a common history.
HeRO assesses the best odometry based on the internal state of the odometry algorithms, while our method aims at finding a more objective metric that can choose fairly the best odometry at every timestamp.
Based on HeRO, Palieri \etal~\cite{palieri2020ral} propose a health monitor system for the odometry estimation, which exploits LiDAR-based scan-to-scan and scan-to-map methods. While this method is similar to our approach, we do not need to manually specify the main odometry or submental ones. Our method chooses the most promising odometry by each candidate's performance scores at every timestamp without relying on any prior assumptions or knowledge about the odometry performance.

Different from all the methods mentioned above, we propose an integrated visual-LiDAR odometry system that uses multiple odometry algorithms in parallel, a set of dynamics and kinematic sanity checks that filter out potential estimation failures, and a point cloud Chamfer distance-based criterion, which chooses the most promising pose estimation at each timestamp, independently from the history.

\section{Our Approach}
\label{sec:main}
	
\begin{figure*}[t]
\centering
\includegraphics[width=0.8\linewidth]{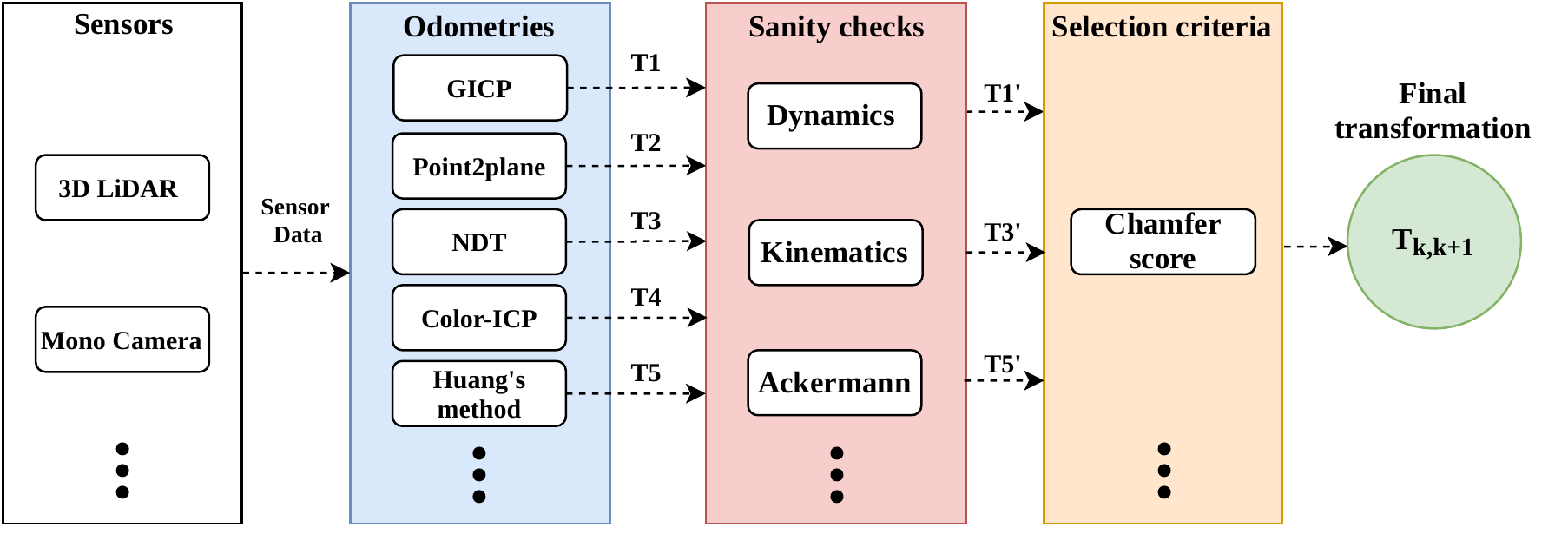}
\caption{System overview. Both 3D LiDAR and camera sensor data are used as the input. We first exploit multiple existing odometry algorithms to process the input data and provide transformation proposals. Those proposals will then be checked by sanity checks based on dynamics and kinematics capabilities. Only those transformations which are labeled as success cases will be evaluated in the next step. The point-to-point Chamfer distance is then calculated to score those success cases. As the result, the best transformation in terms of the calculated scores will be chosen as the final estimation. The final transformation will be saved to the odometry history and employed as an initial guess for the next iteration. Our system is flexible, every submodule could be extended by extra data, methods, or other criteria.}
\label{fig:overview}
\vspace{-0.4cm}
\end{figure*}

We propose a redundant odometry system enhanced with sanity checks. These steps are visualized in~\figref{fig:overview}.
We use the modular design to keep our system flexible.
There are four modules in our system. First, the input preprocessing module exploiting multiple sensing modalities as the input to the system (see~\secref{sec:approach-input}). Second,
the odometry candidates module employing multiple different odometry algorithms to estimate the candidate transformations at each timestamp as odometry proposals (see~\secref{sec:odometries}). Third, the sanity-check module using vehicle dynamics and kinematics properties to check for failure cases in each odometry proposal (see~\secref{sec:sanity-check}). Fourth, the scoring module exploiting the Chamfer distance to score the odometry proposals and selects the final transformation from the most recent vehicle motion (see~\secref{sec:criteria}).

\subsection{Input Preprocessing Module}
\label{sec:approach-input}

In this work, we do not focus on choosing the best sensor or combination to improve odometry but provide a flexible system that allows us to use multiple sensing modalities and ego-motion algorithms in parallel to improve the robustness of the odometry. 
To this end, our system uses the data from a 3D LiDAR scanner and a monocular camera as an example to show that our system can exploit the redundancy of multiple sensing modalities to improve its robustness.
Specifically, we use SIFT features~\cite{tardif2008iros} extract from images obtained by a monocular camera and voxelized point clouds from LiDAR scans as the input to different odometry methods.
Such a redundancy design at the input level has the advantage that our approach is still functional even in case of sensor failures.

\subsection{Stack of Odometry Pipelines}
\label{sec:odometries}
Our approach combines multiple odometry methods into a redundant system, which is robust to failure cases and provides better odometry results.
The odometry algorithms used in this paper are generalized iterative closest point (GICP)~\cite{segal2009rss}, point-to-plane iterative closest point (P2P-ICP)~\cite{rusinkiewicz2001dim}, normal distributions transform (NDT)~\cite{bieber2003iros}, Color iterative closest point (ColorICP)~\cite{park2017iccv}, and Huang's method~\cite{huang2018iros} for combining LiDAR and camera data.
Moreover, we also employ the common constant velocity model (CVM), which directly uses the last timestamp estimation without using the sensor data.
Our approach is not limited only to those methods and can be easily extended.

P2P-ICP~\cite{rusinkiewicz2001dim, zhou2018arxiv} minimizes the sum of the squared distance between the point from the source points and the tangent plane at its corresponding point in the target point cloud for every point.

GICP~\cite{segal2009rss, koide2020easychair} combines the Iterative Closest Point (ICP) and point-to-plane ICP into a single probabilistic framework. It models the locally planar surface structure from both scans and minimizes the objective function in a typical ICP-based framework. This procedure may be interpreted as a plane-to-plane approach embedded in the ICP framework.

ColorICP~\cite{park2017iccv, zhou2018arxiv} optimizes jointly geometrical information (in the sense of point-to-plane ICP) together with RGB color information that is assigned to the LiDAR points which are inside the field of view of the camera(s).

NDT~\cite{bieber2003iros, koide2020easychair} uses a representation of the target point cloud to obtain a smooth surface described as a set of local probability density functions in the subdivided space as a grid of cells. A density function is computed for each cell based on the point distribution within the cell. To estimate the pose, NDT maximizes the likelihood that the current scan points lay on the reference scan surface.

Huang's method~\cite{huang2018iros} employs the 5-point algorithm~\cite{nister2004pami} within a RANSAC loop that finds 5 of the 6 transformation parameters (rotation and translation vector up to scale) between the previous camera image and the current image with the SIFT descriptor. It then uses a grid search-based 1D ICP to estimate the scale.

Besides the odometry information provided by different algorithms based on different sensing modalities, our system also employs the so-called constant velocity model (CVM), which takes the same transformation from the previous timestamp as an additional odometry estimation with the assumption that the acceleration of the car is zero. 
Although the constant velocity model is simple, it will always pass the sanity checks and make the system robust.

Those odometry methods are run in parallel using the same input data and share the same odometry history. 
This means only the transformation that is finally selected by our approach at each timestamp will be stored and serves as the initial guess for all the odometry methods in the next timestamp.

In this paper, we exploit six different odometry methods to exploit the redundancy of multiple odometries to improve overall performance and robustness. 
Note that we can easily add more methods if desired. Since our method can switch dynamically for each timestamp independently from previous decisions, multiple odometry methods can be easily integrated into our system, even during online processing.

\subsection{Sanity Checks}
\label{sec:sanity-check}

In robotics, we typically know the used hardware and the physical model of our robot or a typical car. Therefore, we use the dynamic and kinematic constraints of a car to check the odometry proposals provided by different algorithms, which we refer to as sanity checks.
In this paper, we exploit two types of sanity checks, one based on the dynamic constraint, and the other based on the kinematic constraint.

We refer to the maximum acceleration~$a_{max}$ a typical car can reach for formulating the dynamic constraint. 
Our system first checks the acceleration resulting from all odometry proposals. If the estimated transformation exceeds the bound of the maximum possible acceleration, it will be rejected from further considerations.

\begin{figure}
\centering
\includegraphics[width=0.5\linewidth]{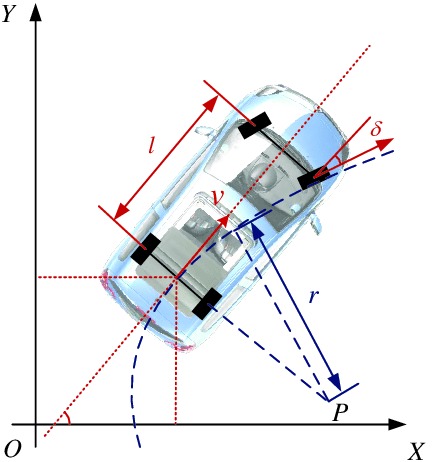}
\caption{Ackermann vehicle model. The variable $l$ is the distance between the front and rear axis, $v$ is the velocity of the rear axis center, $\delta$ is the steering angle, $P$ is the instantaneous center of rotation, and $r$ is the radius around $P$. See~\cite{zong2018sensors} for details.}
\label{fig:ackermann}
\vspace{-0.3cm}
\end{figure}

For the kinematic constraint, we use the Ackermann vehicle model~\cite{zong2018sensors}, which puts constraints on the kinematic possibilities of the car movement. If using a differential drive model, one obviously must use the corresponding model.
As shown in~\figref{fig:ackermann}, given the forward velocity~$v$ and the steering angle rate~$\delta$, the car's trajectory follows the Ackermann model if there is no slip or any other highly dynamic conditions such as drifting or similar. 
It means that in real operations, the car's sideward movements are limited to a certain range.

Applying this fact, our system uses the last pose estimation to calculate the forward velocity. 
Based on the forward velocity and the Ackermann model, the side velocity~$v_\text{s}$ is calculated based on the derivation~\cite{buczko2018ivs},
\begin{align}
	v_{\text{s}}(\dot{\beta}, v) = & f \frac{\frac{v}{f}+l(1-\cos\frac{\dot{\beta}}{f})}{\sin\frac{\dot{\beta}}{f}}
	 (1-\cos\frac{\dot{\beta}}{f})  + l\sin\frac{\dot{\beta}}{f} ,
\end{align}
where $v$ is a forward velocity of the vehicle, and $\dot\beta$ is the turning rate, $f = \frac{1}{\Delta\tau}$, where $\Delta\tau$ is the difference between timestamps of the current frame and previous frame, $l$ is the distance from the rear axis to the placement of the sensor.
A proposal will then be rejected for future considerations if
\begin{equation}
 \abs{v_{s} - v^*} > v_{th},
\end{equation}where $v^*$ is taken from the odometry proposals and $v_{th}$ is a maximum deviation from the side velocity that is acceptable.

Note that even if all sensors or odometry algorithms based on sensor data fail to pass the sanity checks, the constant velocity model will always pass the sanity checks and makes the proposed method fairly robust to failures.

\subsection{Scoring Candidate Transformations}
\label{sec:criteria}

All the odometry proposals, which pass the sanity checks, will be evaluated based on the Chamfer distance metric~\cite{barrow77ijcai} computed on the point cloud. Thus we need at least one sensor that generates point cloud data. We use a 3D LiDAR in our system. The metric on the point cloud is defined as follow,
\begin{equation}
D_{\text{chamfer}}(M,N) = \frac{1}{M} \sum_{m \in M}^{} d_N(m) ,
\end{equation} where $M$~is a set of points in the source point cloud, $N$~is a set of points in the target point cloud, and $d_{N}$~is the minimum distance between point $m$ and its nearest neighbor point in the target point cloud $N$ given the transformation between point cloud $M$ and $N$.
In addition to the fact that point clouds after transformation should close to each other, we apply a small search radius~$r_s$ for correspondence matching and delete the points for which no correspondence could be found within the radius. 
The reason to use the Chamfer distance as the criterion is that it is fast to calculate and independent from all the baseline odometry methods.

Since each odometry candidate in our system estimates the transforms from the same source point cloud to the same target point cloud, this metric allows us to assess the candidate transformations estimated by different odometry methods.
Our system applies this metric to all the success odometry proposals, and the lowest score is chosen as the final odometry result.
\begin{equation}
T^* = \underset{T \in {T_1,T_2,...}}{\arg\min} \frac{1}{M} \sum_{m \in M}^{} d_N(m) ,
\end{equation}
 
The chosen transformation is stored in the odometry history and employed as the initial guess for the next iteration.
We use the last~$N_{\text{map}}$ scans with the stored odometry to build a local point cloud map.

\begin{figure*}
	\hspace*{-0.5cm}
	\vspace{0.2cm}
	\begin{tabular}{ccccc}
		\includegraphics[width=38mm]{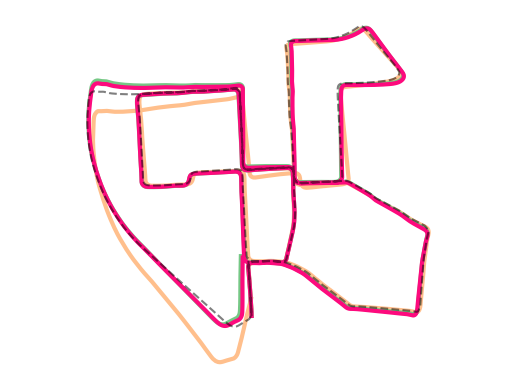} \hspace*{-0.7cm}&   
		\includegraphics[width=38mm]{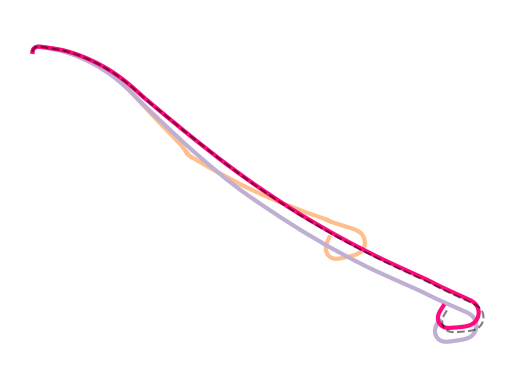} \hspace*{-0.7cm}&
		\includegraphics[width=38mm]{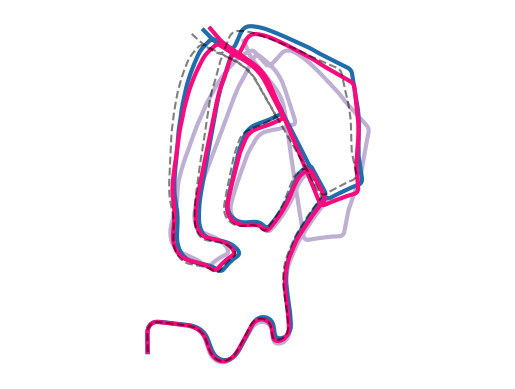} \hspace*{-0.7cm}&  
		\includegraphics[width=38mm]{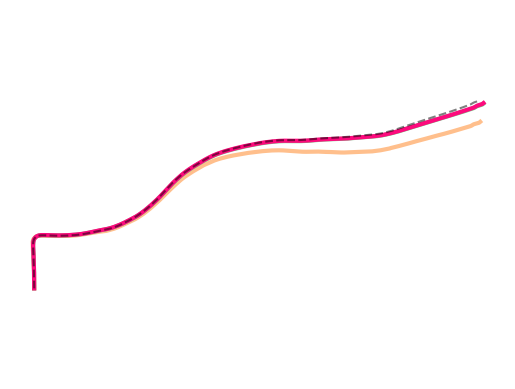} \hspace*{-0.7cm}&  
		\includegraphics[width=38mm]{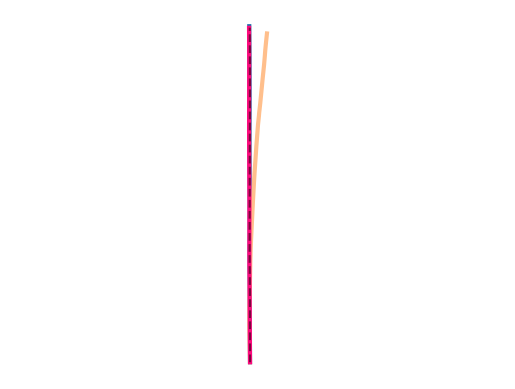} \hspace*{-0.7cm}  \\
		00 \hspace*{-0.7cm} &  
		01 \hspace*{-0.7cm} &  
		02 \hspace*{-0.7cm} & 
		03 \hspace*{-0.7cm} & 
		04 \hspace*{-0.7cm}\\[6pt]
		\includegraphics[width=38mm]{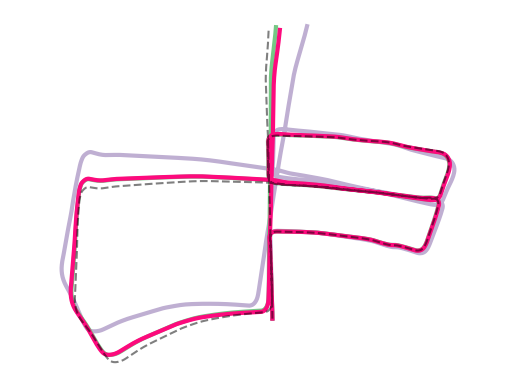} \hspace*{-0.7cm}&   
		\includegraphics[width=38mm]{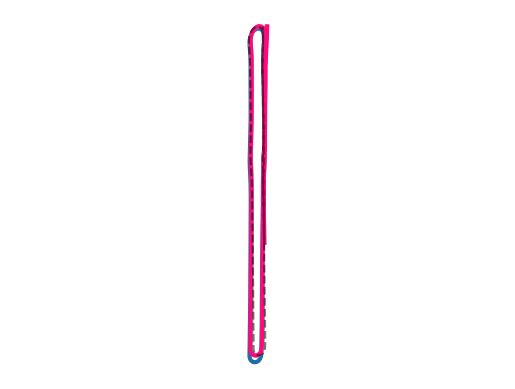} \hspace*{-0.7cm}&
		\includegraphics[width=38mm]{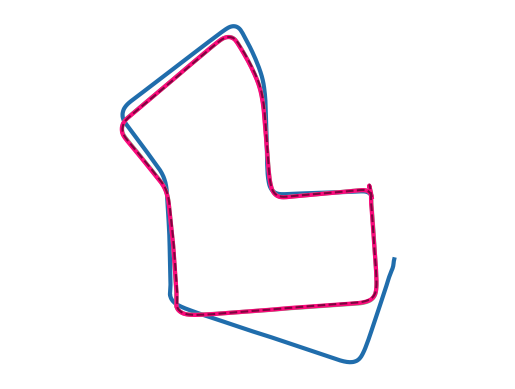} \hspace*{-0.7cm}&  
		\includegraphics[width=38mm]{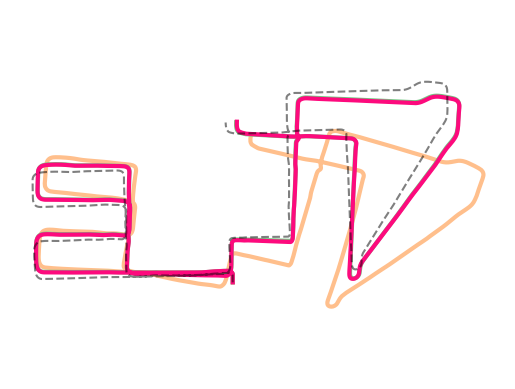} \hspace*{-0.7cm}&  
		\includegraphics[width=38mm]{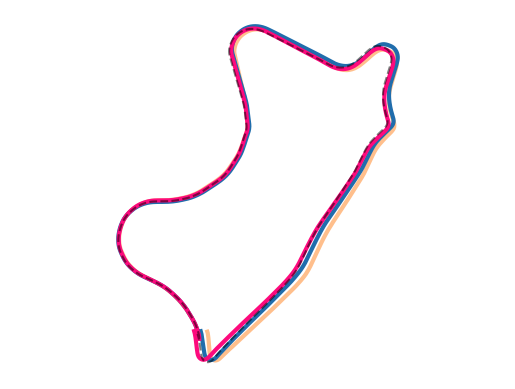} \hspace*{-0.7cm} \\
		05 \hspace*{-0.7cm}&  
		06 \hspace*{-0.7cm}&  
		07 \hspace*{-0.7cm}&  
		08 \hspace*{-0.7cm}&  
		09 \hspace*{-0.7cm}\\[6pt]
		\multicolumn{5}{c}{\includegraphics[scale=0.4]{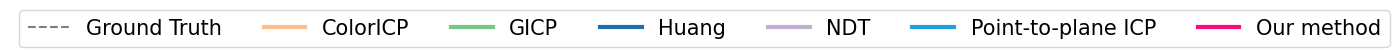} }\\
	\end{tabular}
	\caption{Trajectories of different methods evaluated on the KITTI Dataset. For the sake of clarity, we only show the results of our method together with the best and the worst results of the baseline methods employed by our system.}
	\label{fig:trajectories}
	\vspace{-0.4cm}
\end{figure*}

\section{Experimental Evaluation}
\label{sec:exp}
We present our experiments to showcase the capabilities of our method and to
support our key claims that our approach is able to 
(i)~use redundant odometry pipelines to provide better odometry results,
(ii)~generally yields the best results compared to the individual methods used, and, 	
(iii)~avoids several failure cases in different situations.
For evaluation purposes, we use the KITTI Odometry Benchmark~\cite{geiger2012cvpr} which provides point clouds generated by a Velodyne HDL-64E 3D laser scanner ($10$\,Hz, 64 laser beams, range: 100m) and color images from PointGrey Flea2 video cameras ($10$\,Hz, resolution: $1382 \times 512$ pixels).
We tested our approach on an Intel Core i9-9900K CPU with 16 cores @ 3.60GHz with 32 GB RAM.
For all the experiments, we use $l = 1.0\,m $, $a_{\text{max}} = 6.0\,m/s^2$ and $v_{th} = 0.8\,m/s$, the search radius~$r_s = 0.5\,m$, and the number of scans used to build the local map is $N_{\text{map}} = 10$. 

\subsection{Odometry Performance}

\begin{figure}[t]
	\centering
	\includegraphics[width=0.80\linewidth]{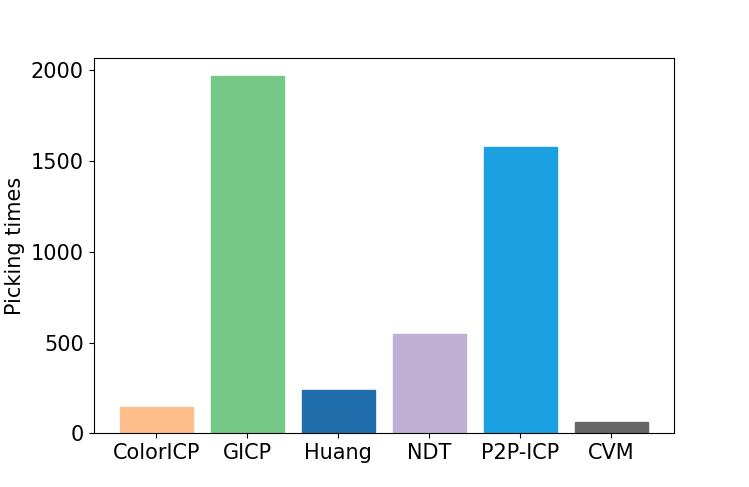}\\
	\vspace{-0.1cm}
	\hspace{0.1cm}
	\includegraphics[width=0.8\linewidth]{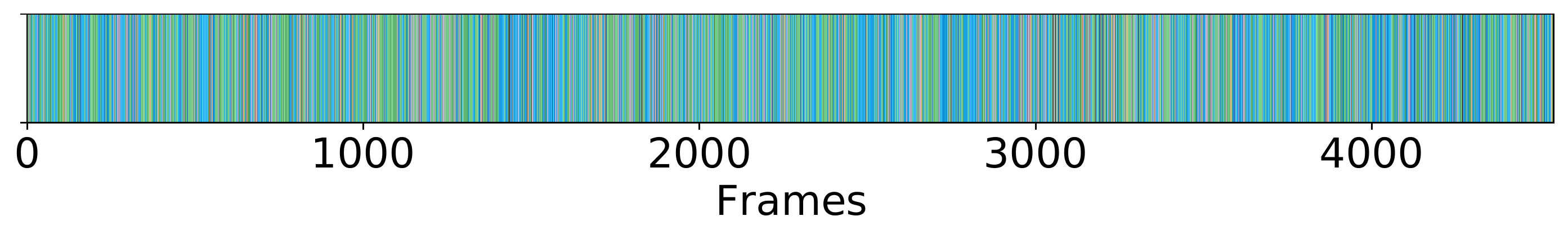}	
	\caption{Frequency of choosing different baseline methods in KITTI sequence 00. The upper figure shows the picking times of different methods, and the lower figure shows the choice of each frame.}
	\label{fig:frequency}
	\vspace{-0.3cm}
\end{figure}

The experiments presented in this section are designed to show our approach's odometry performance and support the claim that using our redundant system yields better overall odometry results.

To evaluate the odometry results, we use the KITTI odometry dataset sequences 00-10 and the KITTI odometry metric~\cite{geiger2012cvpr}, which considers relative errors concerning translation and rotation averaged over different distances.

For the qualitative results, we show the estimated trajectories of our method compared to those generated by the individual odometry algorithms employed by our redundant system, see in~\figref{fig:trajectories}.
For the sake of clarity, we only show the trajectories generated by our method as well as those of the best and the worst individual odometry.
We can see that, in all sequences, our method always tracks the ground truth poses well no matter how bad the worst individual odometry candidate is, and there is no individual odometry baseline that can always generate good results.
   
For the quantitative results, we show all evaluation results on all sequences in~\tabref{tab:kitti-odometry}. 
In terms of translation error, our method is clearly superior in 10 out of 11 sequences to all other methods individually and in the other one sequence achieves a comparable performance to the best individual odometry algorithm. 
Overall, our method outperforms all baselines with the smallest average translation error of~$1.11\,\%$.
In terms of rotation error, our method achieves the second-best average rotation error of~$0.0045\,\text{deg}/\text{m}$.

To further insight into how our system chooses different proposals, we illustrate when and how often different methods have been selected on sequence 00 in~\figref{fig:frequency}.
We can see that our method switches between different proposals frequently according to the sanity checks and scoring results.

\begin{table*}[t]
\centering
\vspace{0.2cm}
\caption{Results on KITTI Odometry}
\scalebox{0.95}{
\scriptsize{
\begin{tabular}{lcccccccccccc}
	\toprule
	& \multicolumn{11}{c}{Sequence} \\
	Approach & 00 & 01 & 02 & 03 & 04 & 05 & 06 & 07 & 08 & 09 & 10 & Average \\
	\midrule
	GICP            & 1.28/0.61 & 3.22/0.91 & 1.78/0.63 & 1.18/0.77 & 1.51/0.93 & 1.18/0.60 & 0.87/0.53 & 0.87/0.57 & 1.32/0.56 & 1.55/0.65 & 1.65/0.69 & 1.49/0.62  \\
	\midrule
	Huang's         & 2.40/0.33 & 6.09/0.21 & 1.50/0.25 & 1.38/0.18 & 0.98/0.22 & 2.06/0.42 & 1.43/0.25 & 6.86/4.04 & 2.71/0.29 & 1.37/0.22 & 2.37/0.24 & 2.35/0.38  \\
	\midrule
	P2P-ICP  & 1.57/0.77 & 5.22/1.56 & 1.58/0.58 & 1.98/1.13 & 1.05/0.56 & 1.32/0.76 & 1.12/0.76 & 0.95/0.62 & 1.37/0.76 & 1.48/0.73 & 1.48\textbf{}/0.70 & 1.64/0.75  \\
	\midrule
	ColorICP        & 3.09/1.60 & 26.08/1.68 & 3.17/1.45 & 2.93/1.40 & 3.11/2.00 & 1.74/0.99 & 0.99/0.63 & 1.60/1.51 & 3.14/1.65 & 2.44/1.33 & 4.04/1.66  & 3.89/1.44 \\
	\midrule
	NDT             & 2.88/1.66 & \textbf{2.42}/0.67 & 3.51/1.58 & 2.89/1.89 & 1.63/1.39 & 2.58/1.33 & 1.27/0.67 & 3.38/2.52 & 2.76/1.43 & 2.64/1.16 & 2.91/1.60 & 2.88/1.46  \\     
	\midrule
		Our method            & \textbf{0.96}/0.43 & 2.53/0.67 & \textbf{1.40}/0.50 & \textbf{1.05}/0.67 & \textbf{0.97}/0.61 & \textbf{0.72}/0.35 & \textbf{0.66}/0.36 & \textbf{0.62}/0.45 & \textbf{0.99}/0.39 & \textbf{0.99}/0.44 & \textbf{1.30}/0.54 & \textbf{1.11}/0.45  \\
	\bottomrule
	\multicolumn{13}{p{\linewidth}}{\vspace{0.01cm}Relative errors averaged over trajectories of $100$ to $800$\.m length:  relative translational error in $\%$ / relative rotational error in $\mathrm{degrees}$ per $100$\.m. \newline Bold numbers indicate best performance in terms of translational error.} \\
\end{tabular}
}
}
\label{tab:kitti-odometry}
\vspace{-0.3cm}
\end{table*}

\subsection{Ablation Study on Odometry Candidates}

\begin{table}[t]
\centering
\vspace{0.2cm}\caption{Ablation study on input odometry methods.}
\scalebox{0.85}{
\begin{tabular}{cccccc} 
\toprule 
P2P-ICP   & NDT & ColorICP & Huang's & GICP & Average\\
\midrule
\cmark &  \cmark &   &   &  &   1.23/0.54  \\
\cmark &  \cmark  & \cmark  &  &  &  1.21/0.52  \\
\cmark & \cmark & \cmark & \cmark  & & 1.17/0.50 \\
\cmark & \cmark & \cmark & \cmark  & \cmark & 1.11/0.45 \\
\bottomrule
\multicolumn{6}{p{\linewidth}}{\vspace{0.01cm}Here we use the same KITTI metric as used in~\tabref{tab:kitti-odometry}. \newline We show the average results of different setups evaluated with all sequences.} 
\end{tabular}
}
\label{tab:ablation}
\vspace{-0.3cm}
\end{table}

The second experiment shows the ablation study on different odometry combinations.
We use the same KITTI odometry dataset sequence 00-10 and the same metric as in the first experiment to evaluate the odometry results of different odometry input combinations.
As shown in~\tabref{tab:ablation}, we can see that no matter what combination of the odometry candidate methods is provided, our system can always outperforms the individual odometry methods (see results in~\tabref{tab:kitti-odometry} for the individual numbers per sequence).

\subsection{Robustness}
\label{sec:robustness}
\begin{figure}[t]
	\centering
	\subfigure[Huang's method failure.]
	{\includegraphics[width=0.49\linewidth]{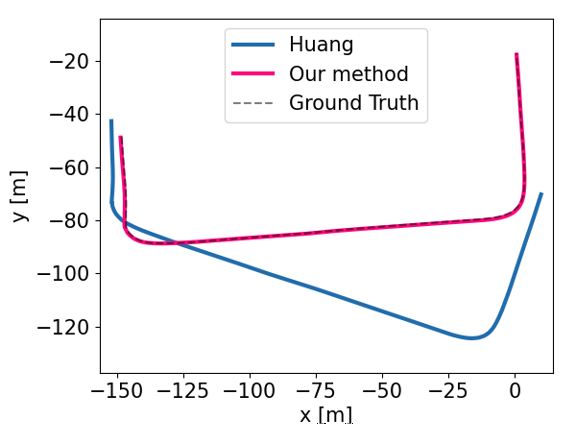}\label{fig:huang_fails}}
	\subfigure[NDT failure.]
	{\includegraphics[width=0.49\linewidth]{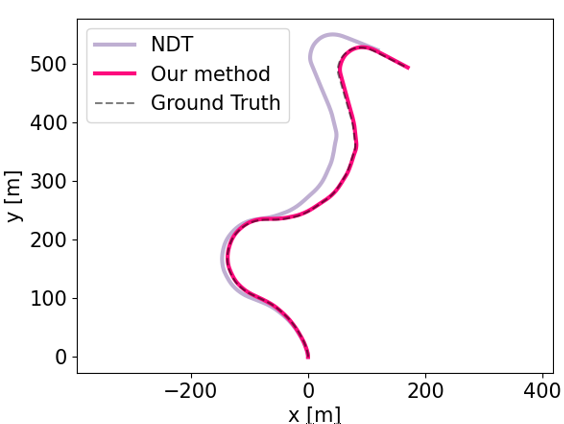}\label{fig:ndt_fails}}
	\subfigure[ColorICP failure.]
	{\includegraphics[width=0.49\linewidth]{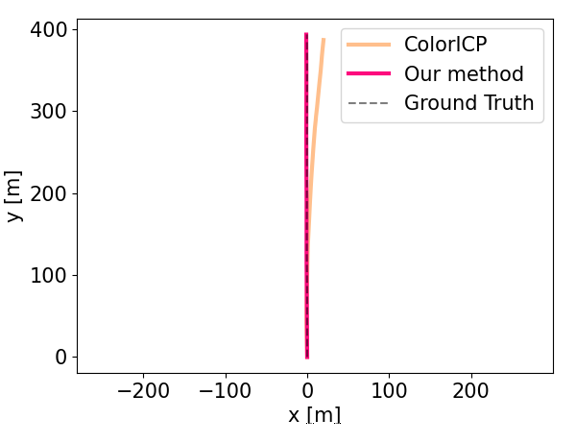}\label{fig:color_fails}}
	\subfigure[Point-to-point ICP failure.]
	{\includegraphics[width=0.49\linewidth]{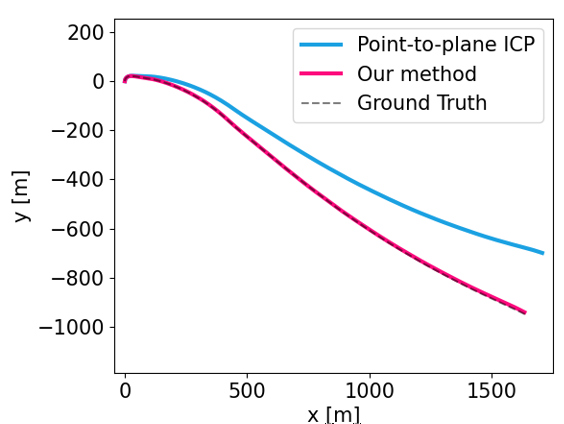}\label{fig:point2plane_fails}}
	\caption{Robustness evaluation of our system. Here we show failure cases of different methods employed by our system (Huang's method, NDT, ColorICP and Point-to-plane ICP respectively). We see that in all failure cases our system work properly.  }
	\label{fig:failures}
	\vspace{-0.3cm}
\end{figure}

The third experiment illustrates our system's robustness and supports the claim that the proposed redundant odometry system is robust to different failure cases.
We use four different cases to show that our system can successfully cover the failure cases where the individual candidate odometry methods fail, see~\figref{fig:failures}. 
The failure cases refer to the situations where the odometry yields a clearly wrong estimate with respect to the ground truth poses.
The failure cases can be caused by the dynamic objects, the lack of features of the environments, or the wrong data associations between different observations. 
In~\figref{fig:failures}, we can see that in all cases when the individual entities fail, our system switches in between the odometry candidates at the right time, choosing the correct estimate and provide proper odometry results.

\begin{figure}[t]
	\centering
	\subfigure[Acc without sanity checks]
	{\includegraphics[width=0.49\linewidth]{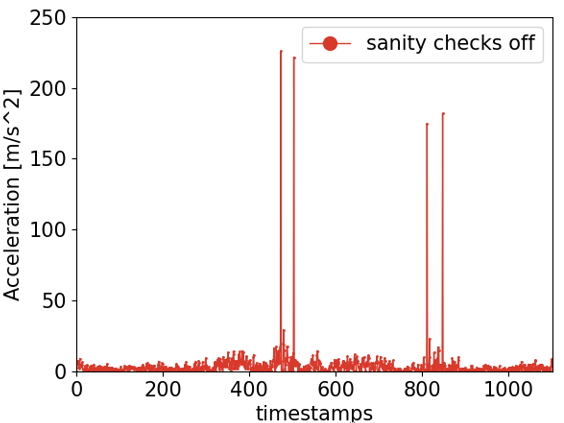}\label{fig:acc_without_limits}}
	\subfigure[Acc with sanity checks]
	{\includegraphics[width=0.49\linewidth]{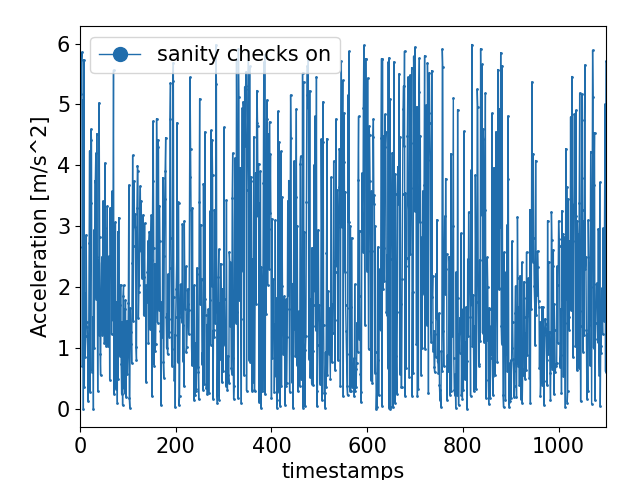}\label{fig:acc_with_limits}}
	\subfigure[Forward velocity comparison.]
	{\includegraphics[width=0.49\linewidth]{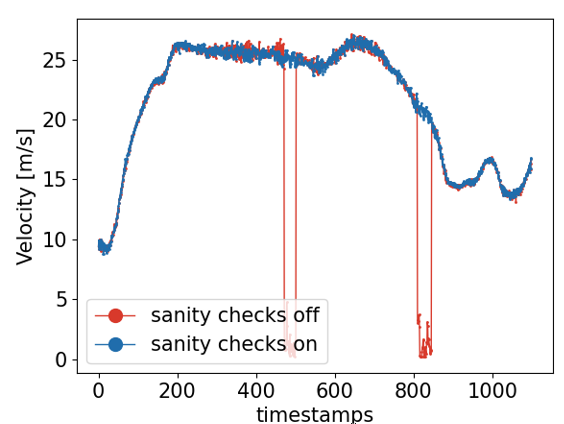}\label{fig:acc_forward_velocity}}
	\subfigure[Trajectory comparison.]
	{\includegraphics[width=0.49\linewidth]{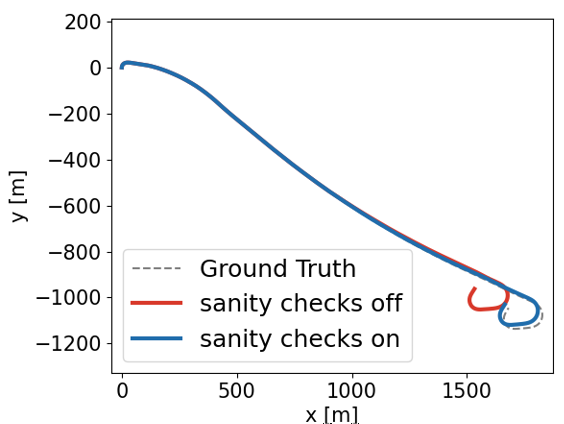}\label{fig:acc_trajectories}}
	\caption{Analysis of sanity checks for our system on dataset 01. (a) shows the noise of acceleration estimation. (b) shows the results after our approach filter out the noise using sanity checks. (c) shows that with sanity checks our method can generate smoother velocity and provide also a better trajectory shown in (d).}
	\label{fig:acceleration_analysis}
	\vspace{-0.4cm}
\end{figure}

To gain further insight, we provide an analysis of the sanity checks. 
If we do not apply sanity checks, our KITTI 00-10 overall score drops from~$1.11\,\%$ to~$1.37\,\%$ in terms of the average translation error.  
The results of the acceleration sanity checks on KITTI sequence 01 is shown in~\figref{fig:acceleration_analysis}.
We can see in~\figref{fig:acc_without_limits} that there are some large accelerations that indicate the failures of the odometry estimations. 
After using sanity checks, see~\figref{fig:acc_with_limits}, our system provides  smoother velocity estimates, see ~\figref{fig:acc_forward_velocity}, and provide better overall results, see~\figref{fig:acc_trajectories}.

\section{Conclusion}
\label{sec:conclusion}

In this paper, we presented a robust and resilient odometry system.
Our method is simple, effective, easy to implement, and we provide the source code for further use. 
Our system exploits the redundancy of running multiple odometry pipelines, including multiple streams, in parallel.
It uses the dynamic and kinematic constraints-based sanity checks and the Chamfer distance-based criteria to avoid failure cases and select the most promising odometry estimation among all proposals. For that, at least one sensor that generates point clouds, such as a LiDAR, is required.
We evaluated our approach on the KITTI dataset and supported all claims made in this paper. 
The experiments suggest that the proposed system is resilient and robust to odometry failures and achieves better performance than all the baselines in terms of odometry accuracy.
Despite these encouraging results, there is space for further improvements. 
In future work, we plan to use learning-based approaches to design better selection criteria for choosing the most promising odometry estimate.


\bibliographystyle{plain}

\bibliography{glorified,new}

\end{document}